\documentclass[nohyperref]{article}
\usepackage{microtype}
\usepackage{graphicx}
\usepackage{subfigure}
\usepackage{booktabs} 

\usepackage{hyperref}

\usepackage[accepted]{icml2022}

\usepackage{amsmath}
\usepackage{amssymb}
\usepackage{mathtools}
\usepackage{amsthm}

\usepackage[capitalize,noabbrev]{cleveref}

\theoremstyle{plain}

\theoremstyle{definition}

\theoremstyle{remark}

\usepackage[textsize=tiny]{todonotes}

\usepackage{tcolorbox}  %

\newcommand{\mynotes}[1]{}

\icmltitlerunning{Bridging AI and Science: Opportunities and Challenges}

\begin{document}

\twocolumn[
\icmltitle{Bridging Machine Learning and Sciences: Opportunities and Challenges}


\centering{\textbf{Taoli Cheng}}

\centering{Mila, University of Montreal}

\centering{chengtaoli.1990@gmail.com}



\icmlkeywords{Machine Learning, ICML}

\vskip 0.3in
]


\begin{abstract}
The application of machine learning in sciences has seen exciting advances in recent years.
As a widely applicable technique, anomaly detection has been long studied in the machine learning community. Especially, deep neural nets-based out-of-distribution detection has made great progress for high-dimensional data. Recently, these techniques have been showing their potential in scientific disciplines. We take a critical look at their applicative prospects including data universality, experimental protocols, model robustness, etc. We discuss examples that display transferable practices and domain-specific challenges simultaneously, providing a starting point for establishing a novel interdisciplinary research paradigm in the near future.
\end{abstract}

\section{Introduction}

The advances in the deep learning revolution have been expanding their influence in many domains and accelerating research in a generalized interdisciplinary manner. Neural nets serving as general function approximators have been employed in scientific applications including object identification/classification, anomaly/novelty detection, autonomous control, and neural net-based simulation, etc. Great successes have been made in multiple scientific disciplines. A typical example is AlphaFold \cite{Jumper2021HighlyAP} for accurate protein structure prediction. At the same time, a lot of progress has been made in physical sciences \cite{Baldi:2014kfa, Ribli:2018kwb}, biology \cite{ching2018opportunities}, molecule generation / drug discovery \cite{Gottipati2020LearningTN},  medical imaging \cite{Zhang2020COVID19SO}, etc.


Despite the successes, there are many challenges or unrecognized pitfalls in transporting machine learning techniques into more traditional science domains. Not being aware of these possible pitfalls could result in vain efforts and sometimes catastrophic consequences in real-world model deployment. 
In the following, we take a holistic look at the current collaborative scheme in machine learning applications for sciences in this new cross-disciplinary research era. The differences between general machine learning (mainly focused on computer vision (CV) and natural language processing (NLP)) and tailored scientific applications reside in all parts of the pipeline.
The following aspects build an intertwined picture in the modern machine learning-assisted scientific discovery:
\begin{itemize}
    \item \textbf{Nature of the data:} In natural sciences especially physical sciences, scientists usually strictly design (and usually simplify) the experimental environments to probe the considered phenomena. Thus the nature of scientific data is under control to be free of external noises due to the laboratory settings. \mynotes{Selective data} (And normally systematic uncertainties can be well estimated through control datasets) However, the format of the data can be more complex (and non-human-readable) compared with natural images.
    \item \textbf{Inference process:}  Model inference in real-world settings can come in complicated and varying formats. As for scientific applications, usually, the experimental focus defines the inference process. And consequently, the inference process affects the result interpretation.
    \item \textbf{Benchmarks:} In the machine learning community usually benchmark datasets used for model evaluation is restricted to a few public datasets such as MNIST \cite{726791}, ImageNet \cite{5206848}, CIFAR \cite{Krizhevsky09learningmultiple}, or SVHN \cite{37648}. This inevitably results in ``overfitted'' strategies and research focuses. In contrast, domain sciences haven't yet built common datasets for model training and evaluation, sometimes resulting in difficulties in model comparison.
    \item \textbf{Uncertainty quantification:} 
    Uncertainty of deep neural net outputs can be hard to quantify due to the complexity involved. When applied in sciences, the capability to incorporate uncertainty estimates in the pipeline is important for rigorous result interpretation and robust model prediction.
    \item \textbf{Generalization and Robustness:} 
    We would like the deep models to perform well at various tasks in real-world environments.
    Furthermore, a common approach in sciences is first training models on simulation data with labeling information. When applied to real data, the dataset shift seeks for remedies to retain accuracy and robustness as in the cases of computer vision or natural language modeling. However, the focus of the simulation-to-real adaptation in sciences differs in the sense that usually the two environments are known however we normally need higher precision.
    \mynotes{\item \textbf{Interpretability/Explanability:} Domain-specific interpretability}
\end{itemize}


Keeping these differences in mind helps shape the research guidelines toward a well-focused and suited technology transfer.
Adapting the workflow according to the needs promotes and secures scientific applications and transforms the research paradigm in a universal manner.
Finally, the interplay between machine learning and scientific discovery benefits from a communal understanding of the field vocabulary, the publishing traditions, the collaboration schemes, and the academic setups \cite{DBLP:journals/corr/abs-1206-4656}.
This new regime solicits novel community infrastructures for more impactful research works in the next few years.

\section{Scientific Discovery}



Modern scientific discovery highly depends on the practices of hypothesis testing and corresponding statistical interpretation, which builds on and renovates the established theories. A recent example of successful scientific discovery is the Higgs boson \cite{PhysRevLett.13.321, Higgs:1964ia} observed at the Large Hadron Collider \cite{ATLAS:2012yve, CMS:2012qbp}. The Higgs boson was predicted by physicists about four decades ago. The properties of the Higgs boson have been well studied during the last decades. This makes dedicated searches possible at the particle colliders.
\mynotes{Another example is the observation of gravitational waves.}

\begin{figure}[htb!]
    \centering
    \includegraphics[width=0.8\columnwidth]{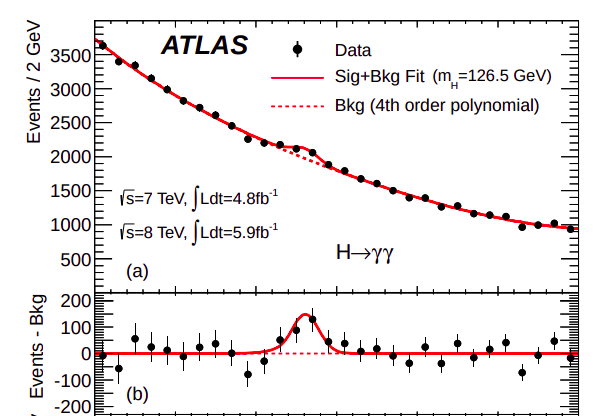}
    \caption{Invariant mass distribution in the channel of Higgs boson decaying to two photons. Image taken from Ref. \cite{ATLAS:2012yve}}.
    \label{fig:higgs}
\end{figure}

The most typical search strategy includes 1) choosing a search channel (i.e., defining the potential observational space for signals), 2) estimating the background using established simulation tools, 3) comparing observed events and the estimated background, and 4) calculating the observed confidence level, setting the observed exclusion limits or claiming an observation under the statistical interpretation \cite{Read:2002hq}. Fig. \ref{fig:higgs} displays a typical discovery histogram for bump-hunt in the dimension of the Higgs boson mass.

\mynotes{bump-hunt https://arxiv.org/pdf/1101.0390.pdf}

\begin{figure}
    \centering
    \includegraphics[width=0.95\columnwidth]{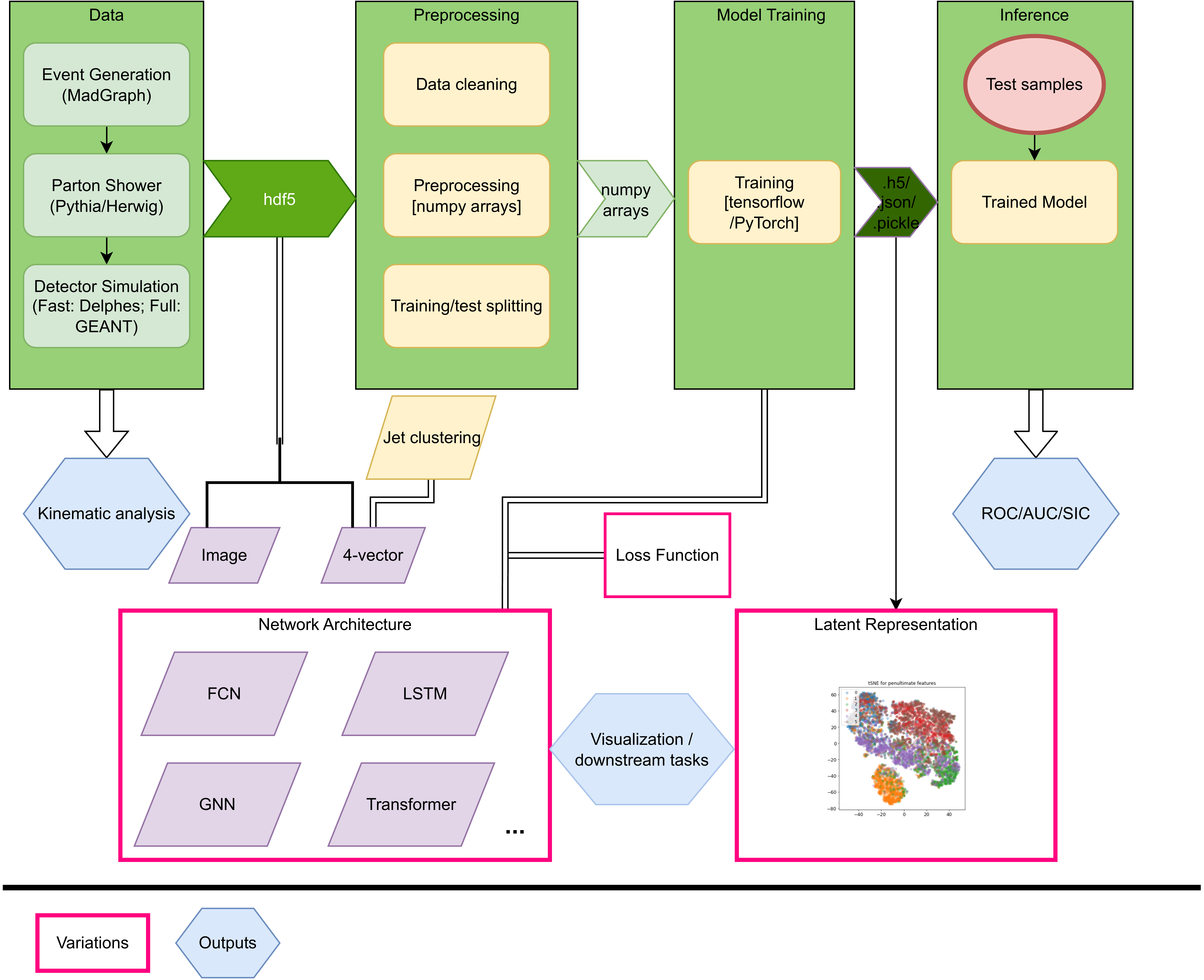}
    \caption{A typical pipeline for machine learning applications in Large Hadron Collider physics. The pipeline is composed of event generation, data preprocessing, model training, and model inference.}
    \label{fig:workflow}
\end{figure}

Translating into the vocabulary of machine learning, the streamlined process consists of 1) defining a (weighted) subset of data classes of interest, 2) training on the datasets under examination, 3) evaluating the trained model on target test datasets, and 4) interpreting the evaluation metrics depending on the context. A typical workflow is depicted in Fig. \ref{fig:workflow}.

The results interpretation highly depends on the associated scientific approaches. Obviously, devoting all the efforts to increasing class-inclusive metrics (as has been widely practiced in the machine learning community) could result in biased research protocols. In the following, we take a careful look at one application area bridging machine learning and sciences: anomaly detection.

\section{An Example: Out-of-Distribution Detection}


Thanks to the capacity to process high-dimensional data, deep neural networks-based out-of-distribution (OOD) detection in computer vision and natural language processing has shown great potential and seen much progress in the past few years \cite{Hendrycks2017ABF, DBLP:journals/corr/abs-2110-06207, Ahmed_Courville_2020, Ren2019LikelihoodRF, Pang2021DeepLF}. Models trained on in-distribution (ID) data are expected to ``know'' what they don't know. Thus they are used to detect unseen patterns as anomalies, by the associated uncertainties or likelihood.
On the other hand, OOD detection also serves as a check for model calibration and robustness against dataset shifts \cite{}. 

At the same time, as an important emerging research area, scientific applications of machine learning are seeing great success in advancing the discovery of novel natural phenomena. Anomaly detection techniques have been used to search for (rare) novel particles for high energy physics \mynotes{\cite{}}, galactic activities for astrophysics \mynotes{\cite{}}, and novel molecular mechanisms for molecular physics \mynotes{\cite{}}. 

\begin{figure}[htb!]
    \centering
    \includegraphics[width=0.8\columnwidth]{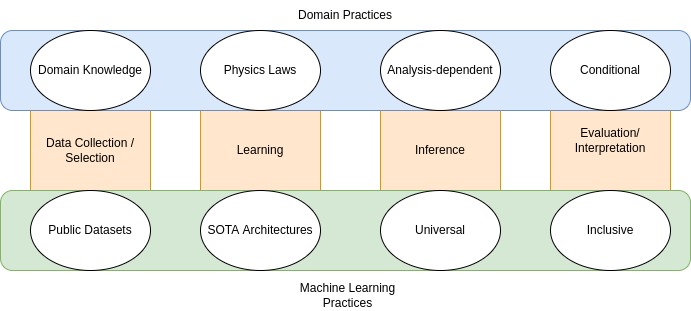}
    \caption{Connection and differences in the pipelines for domain science practices and machine learning practices.}
    \label{fig:schematic}
\end{figure}

Despite these successes, OOD detection in the machine learning community often follows common workflow and conventions, which might result in unexpected failures in real-world applications.
The components of the typical workflow (Fig. \ref{fig:schematic}) reflect fundamental differences and focuses in these two streams of research. The motives and protocols share some common aspects, yet display essential differences and focuses. Understanding these differences in research protocols enables effective application and fosters innovative, impactful collaborations.
We discuss these aspects respectively in the following.

\subsection{Datasets and Input Representation}
\label{subsec:data}

The first and foremost element in the pipeline is the data under consideration.
Computer vision and natural language processing, as the main application areas of machine learning, have restricted input formats. In contrast, data in sciences come in diverse formats, including raw device response records in particle detectors, light spectra received by telescopes, X-ray images, etc. 
\mynotes{On top of that, there are accumulative data v.s. point data, calling for different analysis strategies.}

These differences motivate intriguing research topics specifically for scientific applications. 
On the one hand, the input formats of scientific data request novel tailored neural architectures that are aware of the underlying inductive biases.
For instance, many scientific data have inherent symmetries or are constrained by physics laws or associated invariances. In the same vein, equivariant neural networks \cite{Cohen2016GroupEC} and geometric deep learning \cite{Bronstein2021GeometricDL} have pioneered neural architectures and mechanisms for incorporating the concepts of symmetries. They have been gaining attention in particle physics \cite{Bogatskiy2020LorentzGE}, computational chemistry \cite{Anderson2019CormorantCM, Batzner2022E3equivariantGN}, etc. At the same time, these approaches have boosted research in regular machine learning applications (computer vision, natural language processing, etc.). \mynotes{More citations for this part.}
Similar motives can broaden the research horizon of ``classical'' machine learning, forge multidisciplinary collaborations, and facilitate innovative research advances.
On the other hand, attention-based transformers \cite{vaswani2017attention} have reigned the NLP community and more recently intruded computer vision \cite{dosovitskiy2020image}. They have been successful in processing multiple data formats \cite{reed2022generalist}. These achievements indicate potential directions for cross-disciplinary research protocols with a uniform framework and reusable and shareable modules that can be interfaced with scientific tasks.

Regarding the data collection process, while image labeling is expensive for machine learning research, simulation data with labels in sciences are relatively cheaper to generate (although it could take much computing resources in some cases). That means we can have as much data as we need. As a practical result, some strategies (e.g., those dedicated to small datasets) are not necessarily appropriate for scientific applications. 
On the other hand, models trained on simulated data are confronted with performance degradation-associated distribution shifts when applied to real data. This will be further discussed in section \ref{subsec:robustness}.

\subsection{Inference, Decision, and Evaluation}
\label{subsec:inference}

In most scientific applications, model inference \footnote{We follow the definition in deep learning for the terminology "inference", which mostly refers to the decision-making process at test time.} is closely related to the underlying hypotheses to be tested and thus the evaluation metrics should depend on the chosen test sets which are supposed to resemble the real-world deployment circumstances. In other words, the components of the test sets define the inference strategies. At the same time, the test sets are determined by the science under investigation.

The well-adopted practice in the machine learning community is testing the model on all the classes for a benchmark dataset and reporting class-inclusive metrics \cite{Hendrycks2017ABF}. In most of the research literature, the model evaluation metrics are reported for classifying two large benchmark datasets separately collected (e.g., models trained on CIFAR are tested against other datasets such as SVHN, or even tested on uniform noise as the OOD). This convention is, of course, convenient and effective for a preliminary assessment.

However, in scientific applications, depending on the context, due to the common imbalance of in-distribution classes, it is more suitable to test against class-conditional in-distribution data.
A generic example is that background events have different class weights, sometimes in extreme imbalance. If the phenomenon of interest is detected under the circumstance of a specific set of background events, we will need to adjust our evaluation metrics to be calculated for the distribution of focus. 


\paragraph{Test datasets}
One problem of current scientific applications is the lack of consensus on benchmark test datasets. Researchers might work under different selected subsets, different data qualities, and different simulation settings. \mynotes{This is due to the different working habits and targets of science. Compared with machine learning researchers focused on model improvement, scientists are more focused on the mechanism, working performance, and specific technical settings.} 
Even in more focused sub-fields, researchers usually report model metrics on their own datasets. This brings variations and concerns in model comparison. 

From another practical point of view, robustness examination and stress tests also benefit from a ``complete'' set of test sets. Research schemes might bias toward specific scenarios if they are only tested on a limited set of data. 

\paragraph{Evaluation metrics}
 
For anomaly detection, a widely-reported metric is the Area Under the ROC curve (AUC) score based on the binary ID/OOD classification. 
OOD classes are usually from other datasets different from the training datasets. The selection of ID and OOD datasets is rotating around common benchmarks MNIST/Fasion-MNIST or CIFAR-10/CIFAR-100/ImageNet/SVHN. 
However, this kind of inclusive measure doesn't give much information on context-sensitive use cases. Meanwhile, different datasets have intrinsic dataset shifts which affect the validity of this approach.
Though there are recent proposals \cite{Ahmed_Courville_2020} on transforming the research scheme to a more specific within-benchmark evaluation, we haven't seen large-scale changes in the ML community.

In scientific applications, novel events or phenomena are usually further analyzed by choosing promising working points. One example of useful metrics is the \emph{Significance Improvement Characteristic} \cite{Gallicchio:2010dq} 
$\texttt{SIC} \equiv \frac{\epsilon_S}{\sqrt{\epsilon_B}}$, composed of the background rejection rate $\frac{1}{\epsilon_B}$ ($\epsilon_B$ is equivalently the \emph{false positive rate}) and the signal efficiency $\epsilon_S$ (\emph{true positive rate}). Moving away from the single-metric regime will enrich our understanding of the model performance and allow more productive community-wide research protocols.


\paragraph{Recasting ``OOD Detection / Anomaly Detection"}
Following the previous discussion, it's worth rethinking the terminology "OOD Detection". We argue that we should develop context-aware OOD detection strategies instead of inclusive benchmarks and metrics. \mynotes{one counter-argument is then how we can execute model comparison?}
Real-world applications come in various formats, settings, and objectives. These complexities might make straightforward model comparison difficult. This then leads to what should be the focus of reporting a model's performance and properties. While simple metrics are limited in their representative scope, the underlying rules and mechanisms can be shareable and universal.

\subsection{Model Examination and Failure Analysis}

Failure mode analysis can serve as a portal for better understanding the methods proposed. It's not a ``bonus'' compared with the AUC. Rather, it helps reveal crucial aspects of the model learning mechanisms and provides insights into the robustness under variable deployment circumstances. These inspections often lead to further research ideas and model improvement directions. For instance, OOD detection in computer vision might fail because of the pixel correlation \cite{Ren2019LikelihoodRF}. This inspection leads to the remedies proposed there.

On the other hand, scientific data may have a complex nature that makes model examination difficult compared with the cases in CV or NLP. Usually, they come in formats not ``human-readable''. To interpret model performance and examine model failures would require extra interfaces or tools. Sometimes, involving theoretical and expert-designed features becomes necessary. 

\paragraph{Model Interpretability}
Usually, science domains can have different insights (or theoretical foundations) in interpreting what the corresponding model learns. For images, people peek into textures, shapes, and patterns \cite{Arrieta2019ExplainableAI, 10.5555/3295222.3295230}. But for scientific data, we need to examine the information with equations and physics laws. This brings challenges as well as opportunities for new directions in domain-specific model interpretability.

To better understand what the model has learned, and thus has control over the usage and further development of the algorithms, latent representations can be investigated with the assistance of relevant theories \cite{Cheng:2019isq, Faucett_2021}. These studies open the door for connecting model performance, model understanding, and model robustness.

\subsection{Generalization and Robustness}
\label{subsec:robustness}

Building generalizable and robust models is the goal of every machine learning practitioner.
There are different understandings of what “robustness” \cite{2021arXiv211215188B} stands for, especially across different domains. In the ML community, adversarial attacks \cite{DBLP:journals/corr/GoodfellowSS14} caused by perturbations in the input space are under frequent scrutiny.
Models robust to malicious attacks have been developed and investigated. 
In scientific applications, we may have a different definition and focus regarding model “robustness”. In anomalous signal detection, we would like the model to 1) be effective across an extensive range of unseen signals, 2) have less distortion under the simulation-to-data shift, and 3) be invariant to spurious correlations.

\paragraph{Uncertainty Quantification and Model Calibration}

In counting experiments, which is the main probing method in particle physics, statistical and systematic uncertainties \cite{1989sgtu.book, Cranmer:2014lly, Demortier:1099967} are the common uncertainties we consider for interpreting the results. While the statistical uncertainty is simply determined by the intrinsic randomness of the data itself, the systematic uncertainty generally consists of different factors in the experimental pipeline, including the object reconstruction efficiency, the simulation/modeling uncertainty, etc.
 Uncertainty plays an important role in quantifying the signal significance and credibility of the observation. An example is the 5-sigma standard for claiming an observation, which means only with a probability of about one in a million the observation is due to random fluctuations. Larger uncertainties lead to less precise measurements and less effective signal detection.

\mynotes{Uncertainty in deep learning models: an additional associated score presenting the uncertainty. Especially, when a trained classifier is confronted with out-of-distribution data, we would like to the model to be able to output information on ...
There is no straightforward method for representing the predictive uncertainties of deep models.
}

Inserting neural nets into the pipeline could result in complicated uncertainty quantification. Calibration of the outputs and estimation of the effects of nuisance parameters need extra effort.
Meanwhile, the terminologies are different in the machine learning community.
Uncertainties in deep neural models are usually categorized into \emph{epistemic uncertainty} (model uncertainty) and \emph{aleatoric uncertainty} (data uncertainty) \cite{Gal2016UncertaintyID}.
At first sight, the \emph{aleatoric uncertainty} seems identical to the \emph{statistical uncertainty}, while the \emph{epistemic uncertainty} shares a similar meaning with the \emph{systematic uncertainty}. 

Quantifying what the neural networks don't know, or the uncertainty of the outputs, is important to ensure we have a trustworthy deployment of the models \cite{Ovadia2019CanYT, Malinin2018PredictiveUE}.
Calibration is one method to quantify the quality of model uncertainty.
(A calibrated model should have a decent alignment between the model confidence and the actual likelihood.)
Especially for OOD detection, out-of-distribution uncertainty (“know what is unknown by the model”) is also an important indicator. One typical method that has a long history is ensemble models \cite{NIPS2017_9ef2ed4b}. At the same time, in the framework of Bayesian Neural Networks, the Monte Carlo drop-out \cite{gal2016dropout} technique can serve as a surrogate for estimating model uncertainty. 

Meanwhile, taking epistemic uncertainty into account can increase model robustness and improve OOD detection performance \cite{NIPS2017_9ef2ed4b}.
Correspondingly in scientific applications, systematic uncertainty has been associated with nuisance parameters \cite{Dorigo:2020ldg, dAgnolo:2021aun, Ghosh:2021roe}. And incorporating this uncertainty in the training strategy can result in more robust and powerful models.

A directly related topic is then model calibration.
Model outputs in probabilities can be misaligned with the predicted accuracy. This misalignment might yield erroneous predictions which high confidence, resulting in failures in a real-world deployment.
Model uncertainty calibration \cite{10.5555/3305381.3305518, Gal2016UncertaintyID} has been revisited for deep neural networks.
Correspondingly, approximate likelihood ratio \cite{Cranmer:2015bka} via neural classifiers can be combined with output calibration to estimate likelihood ratios in precision. 

\paragraph{Distribution Shift}
Due to the large dimensionality under consideration in deep learning models, the effects under distribution shift are hard to quantify. Robustness under distribution shift can be realized through adversarial training strategies \cite{Ganin2016DomainAdversarialTO, Li2018DomainGV}. Other approaches \cite{magliacane2018domain} leverage domain invariances to achieve robustness. Similar approaches have been taken in High Energy Physics \cite{Louppe2017LearningTP}, though used in another context.

\mynotes{Background estimation: simulation-based vs data-driven (Sideband, ABCD, matrix method.)}

In the same vein, for many scientific domains, simulation plays an important role in modeling the background. Especially for supervised learning, data with label information are mostly dependent on the simulator.  Models trained on simulation data will have degraded performance when directly applied to real data. Though there are workarounds for background estimation, achieving domain invariance brings more opportunities and makes even complex search strategies possible.

\paragraph{Spurious Correlation}

A critical factor affecting OOD detection performance and the applicational potential is the spurious correlations \cite{McCoy2019RightFT, Ming2021OnTI} -- the correlations might not hold further in unseen datasets. Manifest in computer vision, for instance, the active pixel number might be a strong confounder for OOD detection in the generative approach \cite{NEURIPS2019_1e795968}.
They may result in defective model performance \mynotes{\cite{}}, severe failure of model deployment (under distribution shift), and harmful consequences in extreme cases.

In scientific data, spurious correlations have slightly different origins and underlying contexts. And their application consequences are closely related to the scientific approach (or search strategy). Moreover, our laboratory environment provides possibilities to better control spurious factors.

Cleverly ameliorating spurious correlations could even bring improved OOD detection performance. For instance, incorporating invariances can serve as a strategy for improved representation learning and boosted performance. \mynotes{\cite{Cheng:2022gma}}

\mynotes{
\paragraph{An Example: Mass Decorrelation in Bump-Hunting in HEP}
In resonance search at the Large Hadron Collider, 
}

\paragraph{Invariances}

Spurious correlation, as discussed before, could result in catastrophic  consequences if not dealt with care when the model is deployed in real-world applications.
Incorporating invariances 
or employing invariant training objectives/procedures \cite{Arjovsky2019InvariantRM} to ``environments'' has been systematically discussed in the context of computer vision. 

However, for science data, the related invariances could come from a different source and at the same time can be incorporated in a more disciplined manner. As a concrete example, in searching for a specific signal type (e.g., mass resonance), we expect event excess in that observational dimension (e.g., invariant mass). Incorporating model invariance w.r.t. the observational dimension can serve as a critical handle for effective signal detection and better generalization \cite{Cheng:2022gma}.

\section{Summary}

We're at a pivotal moment in history where advanced computational techniques can drive groundbreaking scientific discoveries. Deep learning as a unifying tool, can tackle diverse scientific problems, facilitating an enriching exchange between scientific domains and machine learning that fuels innovation.
The coming decades call for a novel collaboration scheme of machine learning and sciences, which will foster great opportunities for machine learning researchers and domain scientists. By clarifying the current research protocols, identifying obstacles and pitfalls, discussing possible solutions, and conceiving future directions, we can advance this career further.


In order to make the discussion more concrete, we utilize anomaly detection -- a popular topic in machine learning and an emerging research direction in sciences -- to take a detailed look at the current gaps.
We discussed different components in the pipeline and analyzed the communal practices and the differences in the machine learning community and the science.
We also discussed opportunities and challenges and proposed potential solutions for the new generation's collaborative innovation in this new era.

%


\section*{Acknowledgements}

This work is supported by the IVADO postdoctoral research funding.


\nocite{langley00}

\bibliography{main}
\bibliographystyle{icml2022}

\end{document}